\documentclass[runningheads]{llncs}
\usepackage[T1]{fontenc}
\usepackage{graphicx}
\usepackage{amssymb}
\usepackage{amsmath}
\usepackage[misc]{ifsym}
\usepackage{url}
\usepackage{booktabs}
\graphicspath{ {./fig} }
\usepackage{enumitem}
 \usepackage[normalem]{ulem}

 \usepackage{ctable}
 \useunder{\uline}{\ul}{}
\newcommand{\netname}{\textsc{TimeHetNet}}

\newcommand{\timehetnet}{\textsc{TimeHetNet}}

\newcommand{\graphhetnet}{\textsc{GraphHetNet}}
\newcommand{\graphetnet}{\textsc{GraphHetNet}}
\newcommand{\ght}{\textsc{GHN}}

\begin{document}
    \title{Few-shot human motion prediction for heterogeneous sensors}
    \titlerunning{Few-shot human motion prediction for heterogeneous sensors}

        \author{Rafael Rego Drumond$^*${\Letter}\orcidID{0000-0002-6607-3208}\and
        Lukas Brinkmeyer$^*${\Letter}\orcidID{0000-0001-5754-1746}  \and Lars Schmidt-Thieme \orcidID{0000-0001-5729-6023}}
        \def\thefootnote{*}\footnotetext{Equal contribution}
         \def\thefootnote{}\footnotetext{\textbf{Acknowledgements}: This work was supported by the Federal Ministry for
Economic Affairs and Climate Action (BMWK), Germany, within the framework
of the IIP-Ecosphere project (project number: 01MK20006D).
}
\def\thefootnote{}\footnotetext{\textbf{Accepted at PAKDD 2023}}
        \authorrunning{R.R. Drumond et al. and L. Brinkmeyer}
        
        \institute{University of Hildesheim, Germany 
        \email{$\{$radrumond,brinkmeyer,schmidt-thieme$\}$@ismll.uni-hildesheim.de} }
        
        \tocauthor{Rafael~Rego~Drumond, Lukas~Brinkmeyer, and Lars~Schmidt-Thieme}

     \maketitle \begin{abstract}
Human motion prediction is a complex task as it involves forecasting variables over time on a graph of connected sensors. This is especially true in the case of few-shot learning, where we strive to forecast motion sequences for previously unseen actions based on only a few examples. Despite this, almost all related approaches for few-shot motion prediction do not incorporate the underlying graph, while it is a common component in classical motion prediction. Furthermore, state-of-the-art methods for few-shot motion prediction are restricted to motion tasks with a fixed output space meaning these tasks are all limited to the same sensor graph. In this work, we propose to extend recent works on few-shot time-series forecasting with heterogeneous attributes with graph neural networks to introduce the first few-shot motion approach that explicitly incorporates the spatial graph while also generalizing across motion tasks with heterogeneous sensors. In our experiments on motion tasks with heterogeneous sensors, we demonstrate significant performance improvements with lifts from $10.4\%$ up to $39.3\%$ compared to best state-of-the-art models. Moreover, we show that our model can perform on par with the best approach so far when evaluating on tasks with a fixed output space while maintaining two magnitudes fewer parameters.

\end{abstract}         \section{Introduction} \label{chap:introduction}
    
    Time-series forecasting has become a central problem in machine learning research as most collected industrial data is being recorded over time. A specific application for time-series forecasting approaches is human motion prediction (or human pose forecasting), in which a multivariate time-series is given in the form of a human joint skeleton, and the objective is to forecast motion sequences based on previous observations. This area has recently seen various applications ranging from healthcare \cite{taylor2020intelligent} and smart homes \cite{jalal2019wrist} to robotics \cite{unhelkar2018human,liu2019deep}. Deep learning methods have shown state-of-the-art performances in the task of human motion prediction in recent years with a focus on popular time-series forecasting models including LSTM's and GRU's \cite{martinez2017human}, temporal autoencoders \cite{butepage2017deep}, and more recently transformer-based approaches \cite{mao2020history,mao2021multi}. Moreover, employing graph-based models has shown to be advantageous in cases where the human joint skeleton can be utilized \cite{li2020dynamic}.
    
    \begin{figure}[t] \label{fig:pred} \centering
        \includegraphics[width=0.042\textwidth]{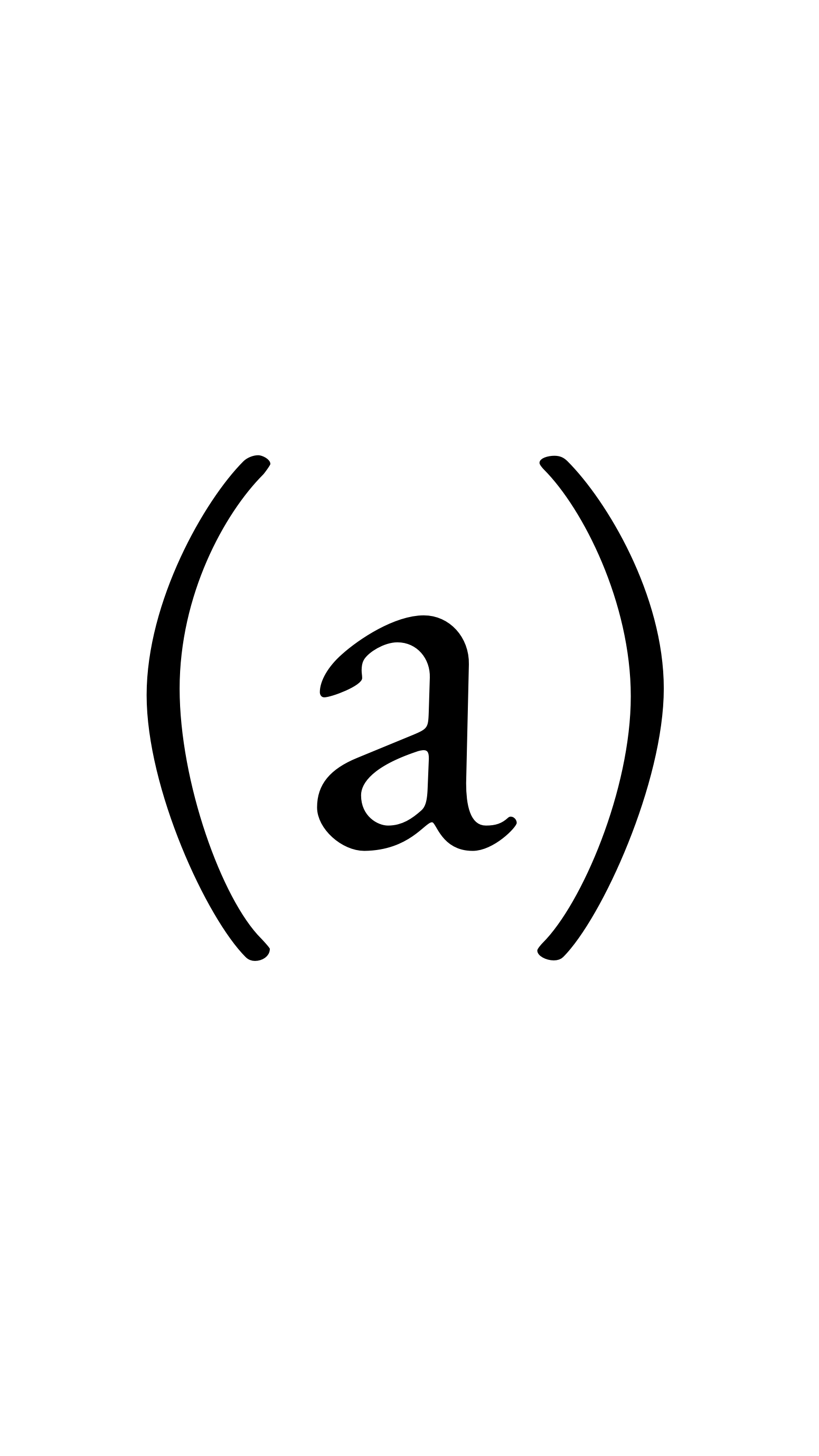} \includegraphics[width=0.91\textwidth]{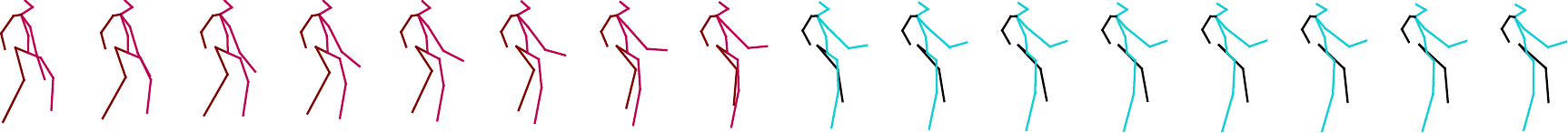}\\
        
        \includegraphics[trim={0.1cm 5cm 0.1cm 1.750cm},clip,width=0.042\textwidth]{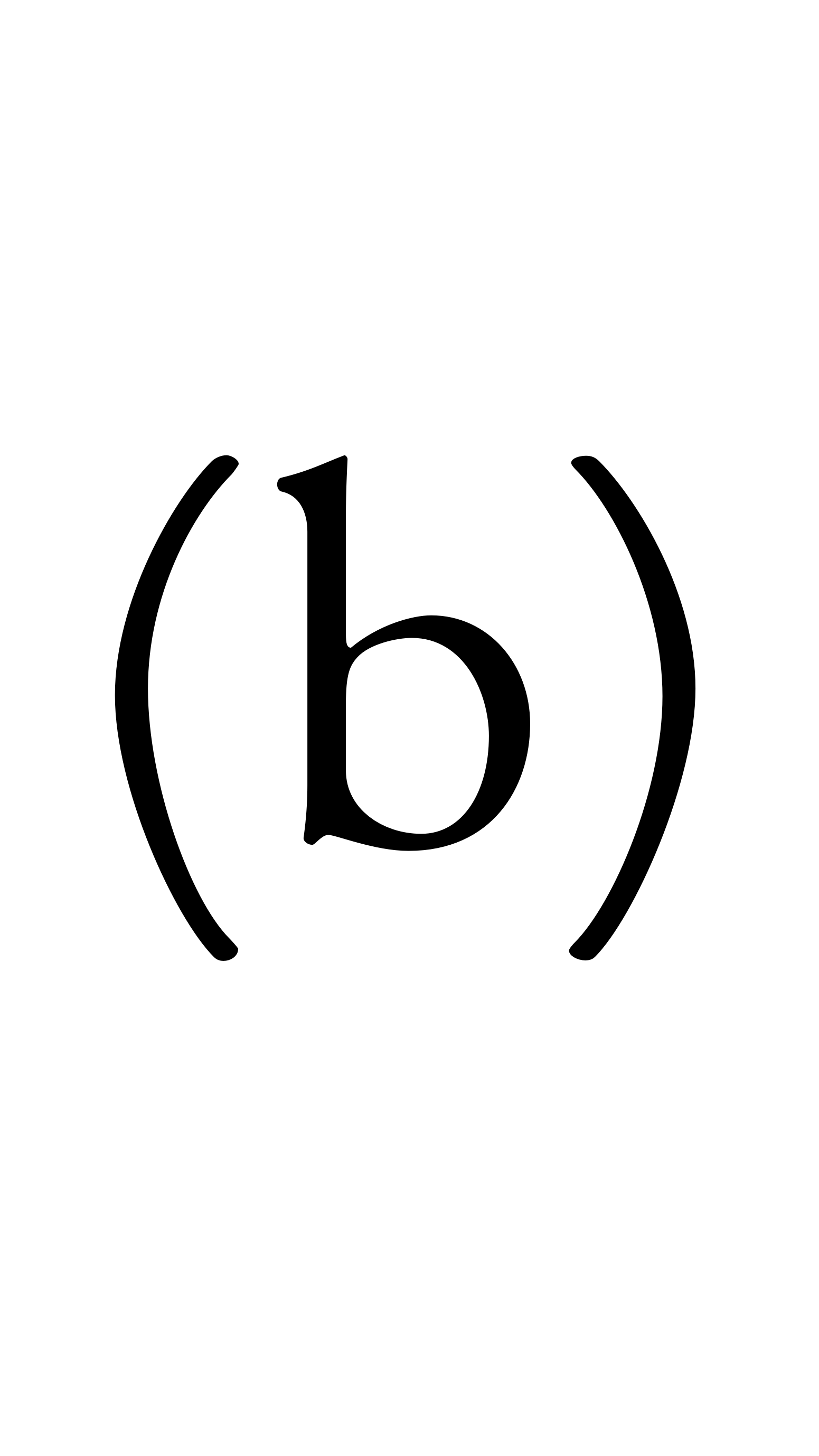} \includegraphics[width=0.91\textwidth]{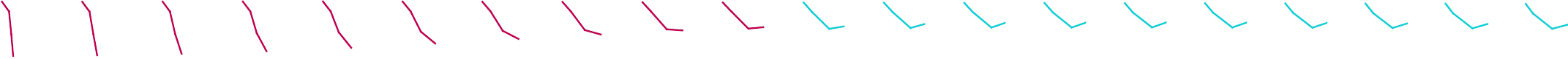}\\
        
        \includegraphics[trim=0.1cm 3.5cm 0.1cm 1cm,clip,width=0.042\textwidth]{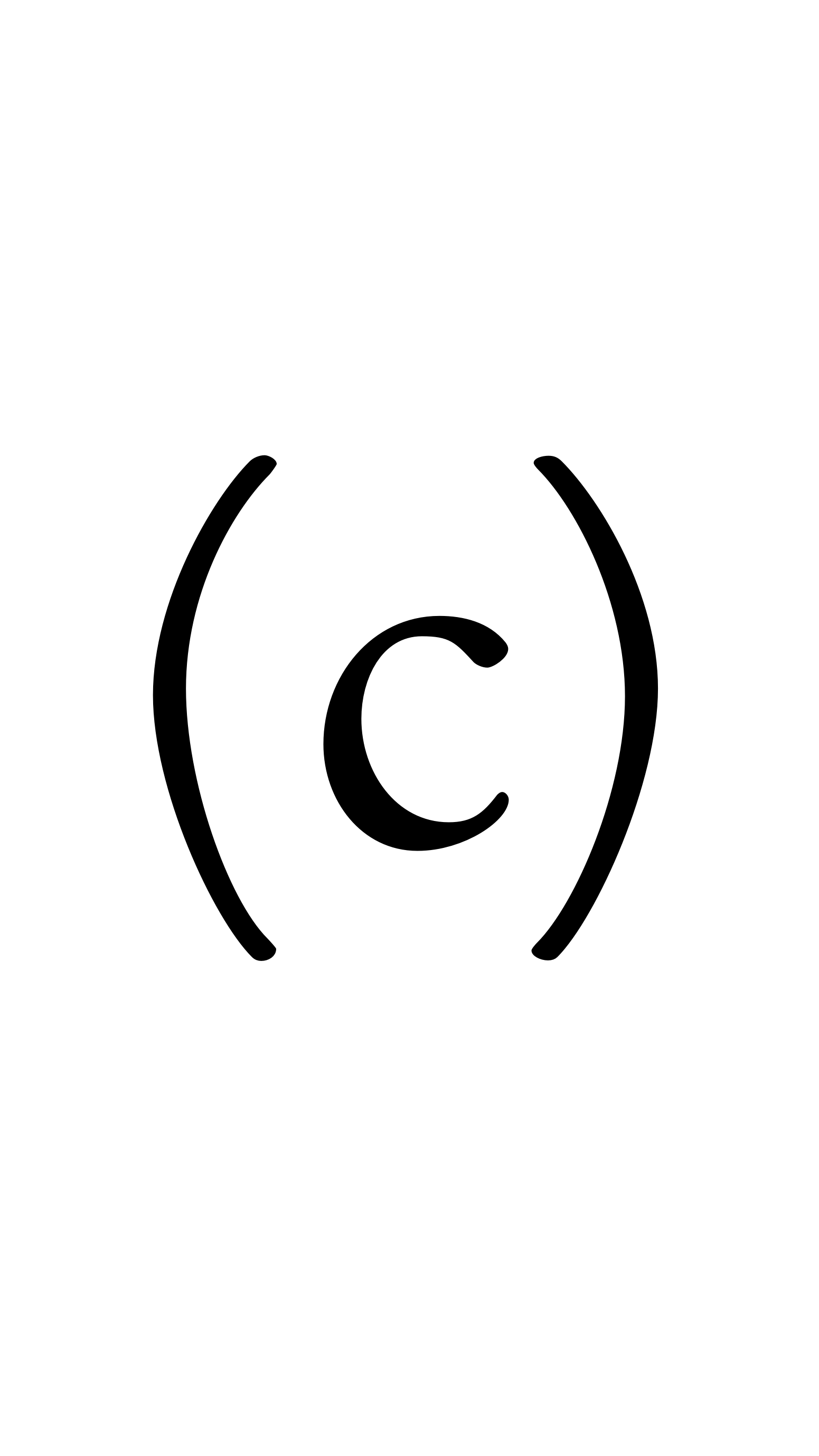} \includegraphics[width=0.91\textwidth]{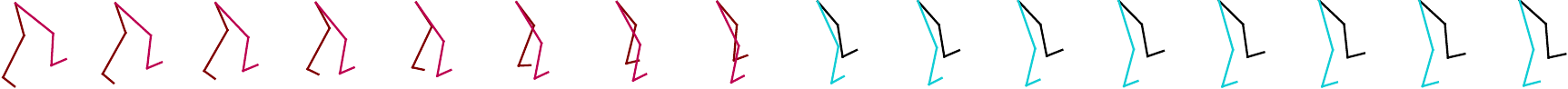}\\     
        
        \caption{Examples of three motion prediction tasks with observations (red) and forecasts (teal). (a) being a standard motion prediction on the full graph, while (b) and (c) are forecasts based on only a subgraph of the sensor skeleton learned by the same model.}
        \label{fig:motion}
    \end{figure}
    
    In few-shot motion prediction, we strive to forecast the motion for previously unseen actions using only a few labeled examples, in contrast to standard human motion prediction, where the training dataset already contains sufficient samples for each action that will be encountered during testing. This can be highly beneficial in practice, as it eliminates the need for such a dataset and allows for a more flexible application. For example, end users can then add new motions by demonstrating an action a few times before the model can accurately classify and forecast future frames. Current approaches for motion prediction are limited to a fixed attribute space such that every observation needs to be recorded across the same set of input sensors. However, an ideal model should be able to cope with only a subset of motion sensors, as not every user should be required to have motion sensors for the full human skeleton. Also, not every action requires information from every possible sensor, e.g., recordings of only the arm for the motion "waving." An example of this is shown in Figure~\ref{fig:motion}, where a motion prediction for the complete human skeleton, but also partial subgraphs of it, is demonstrated. In few-shot learning, this setup is referred to as learning across tasks with heterogeneous attributes \cite{iwata2020meta,brinkmeyer2022few} and is typically tackled by employing a model which operates on attribute sets (in contrast to vectors) which inherently do not possess any order.
    
    In human motion prediction, the attributes represent sensors distributed on a human skeleton \cite{h36m_pami,IonescuSminchisescu11}, meaning they possess order in the form of a graph structure. This information is often used in approaches for classical human motion prediction but not in the current literature for few-shot motion prediction. In a few-shot setting for tasks with heterogeneous sensors, the model would encounter varying graphs in training, similar to classical graph classification approaches \cite{kipf2016semi}. In this chosen scenario, each motion prediction task has a different set of sensors (attributes) that are shared across their subjects, each frame (or pose) corresponds to one time-step, and, finally, the placement of the existing sensors on the subject's body is represented by the task's graph.
    In this work, we propose the first model for few-shot motion prediction that incorporates the underlying graph information while generalizing across tasks with heterogeneous sensors. We evaluate our approach on different variations of the popular Human3.6M dataset and demonstrate improvements over all related methods.
    The contributions of this work are the following:
    
    \begin{enumerate}
        \item We propose the first model for few-shot motion prediction that incorporates the underlying graph structure, while also being the first model for few-shot motion prediction which generalizes to motion tasks with heterogeneous sensors.
        \item We conduct the first few-shot human motion experiments on tasks with heterogeneous sensors where we can show significant performance improvements over all related baselines with performance lifts ranging from $10.4\%$ to $39.3\%$.
        \item We demonstrate minor performance improvements over state-of-the-art approaches in the standard experimental setup while maintaining two magnitudes fewer parameters within our model.
	    \item We also provide code for our method as well as for two of our baselines that have not published a working implementation.
    \end{enumerate}     \section{Related Work} \label{chap:relatedwork}

This work lies in the intersection of few-shot learning (FSL) and human motion prediction. Thus we will discuss the related work of both areas before summarizing the work in the analyzed field. FSL \cite{wang2020generalizing} aims to achieve a good generalization on a novel task that contains only a few labeled samples based on a large meta-dataset of related tasks. There are different techniques, including metric-based \cite{snell2017prototypical}, gradient-based \cite{finn2017model}, and memory-based approaches \cite{yoon2019tapnet}, that have shown successful results. They typically all involve meta-training across the meta-dataset while performing some adaptation to the test task at hand.

Recently, different works have tried to extend few-shot learning to generalize across tasks that vary in their input \cite{brinkmeyer2019chameleon} or output space \cite{drumond2020hidra}. One is to apply permutation-invariant and -equivariant models that operate on sets of elements through the use of deep sets \cite{zaheer2017deep}. \netname{} \cite{brinkmeyer2022few} extended this approach to perform few-shot time-series forecasting on tasks with a single target variable and a varying amount of covariates. \textsc{chameleon} \cite{brinkmeyer2019chameleon} allows vector data-based tasks to have different shapes and semantics as long as the attributes can be mapped to a common alignment. All these methods, however, did not consider any structural relation between the attributes and operate purely on sets of scalar attributes.

Motion Forecasting (or Pose Forecasting, or Pose Estimation) is the task of predicting the subsequent frames of a sequence of human poses. This data can be collected directly as images, or with accelerometers and gyroscopes \cite{Parsaeifard_2021_ICCV}. Most approaches naturally rely on standard deep learning methods for time-series forecasting such as Variational Auto Encoders, LSTMs, and recurrent convolution networks \cite{drumond2018peek,taylor2020intelligent,jalal2019wrist,liu2019deep,unhelkar2018human}. These methods are devised for different motion applications that vary in the type of sensors or forecasting length. For example, \textsc{Peek} \cite{drumond2018peek} and the work of Jalal et al. \cite{jalal2019wrist} require only motion data from the arms, while Gui et al. \cite{gui2018few} use the rotation of the main joints of the complete human body to predict future time-steps. None of these approaches, however, are designed to handle tasks where the set of motion sensors varies.

There are two recent approaches published for few-shot human prediction that we will focus on in this paper as baselines. \textsc{paml} \cite{gui2018few} consists of the popular meta-learning approach \textsc{maml} \cite{finn2017model} operating on top of the classical motion prediction model \textsc{residual-sup} \cite{martinez2017human}. It incorporates a simple look-ahead method for the decoder weights based on pre-trained weights on a bigger dataset to fine-tune the model for a new task. \textsc{MoPredNet}~\cite{zang2021few,zang2022few} is a memory-based approach that uses attention and an external memory of pretrained decoder weights to compute the weights for a new task. Although these two methods work with different tasks separated by the human action performed in each pose sequence, they require the same set of sensors for each task.
    
    In this paper, we present \graphhetnet{} (\ght): a graph-based approach to adapt the \timehetnet{} \cite{brinkmeyer2022few} architecture to train across different human motion detection tasks with heterogeneous sensors by integrating information of neighboring sensors through the application of graph convolutional networks \cite{kipf2016semi}. Thus, we can combine both graph and time-series information into our few-shot predictions.

     \section{Methodology} \label{chap:methodology}

\subsection{Problem definition}

We formulate few-shot motion prediction as a multivariate temporal graph problem. In standard human motion prediction, we are given a graph $\mathcal{G}=(\mathcal{V},A)$ as predictor data where the vertex set $\mathcal{V}$ consists of $C$ motion sensors $\{1,...,C\}$ and $A\in \mathbb{R}^{C\times C}$ is a symmetric adjacency matrix representing the edges between sensors with $A_{ij}=1$ iff sensors $i$ and $j$ are connected by an edge, e.g., an elbow and the shoulder. We also refer to this graph as motion graph, as it contains all the motion sensors. Additionally, we are given a set of node features $X=\{x_{ict}\}\in \mathbb{R}^{I\times T \times C}$ which represent a multivariate time-series with $I$ instances over $T$ time steps for the $C$ motion sensors. We want to forecast the next $H$ time steps given the observed $T$ such that our target is given by $Y\in\mathbb{R}^{I\times H\times C}$.\\
Extending this formulation to few-shot learning, we are given a set of $M$ tasks $D:=\{(D_1^s,D_1^q),...,(D_M^s,D_M^q)\}$ called meta-dataset where each task consists of support data $D^s$ and query data $D^q$ with $D^s_m:=(\mathcal{G}_m,X^s_m,Y^s_m)$ and $D^q_m:=(\mathcal{G}_m,X^q_m,Y^q_m)$. The graph is shared across instances of both support and query for a given task. We want to find a model $\phi$ with minimal expected forecasting loss over the query data of all tasks when given the labeled support data, and predictor of the query data:
\begin{equation}
   \min_\phi \frac{1}{M} \sum_{(D^s_m,D^q_m)\in D}  \mathcal{L}(Y^q_m,\phi(G_m,X^q_m,D^s))
\end{equation}
In the standard setting $\mathcal{G}_m=\mathcal{G}_{m'} \ \forall m,m'\in M \ (m\neq m')$, which means that the structure of the graph $\mathcal{G}$ does not vary across the meta-dataset. Thus, each sample of each task contains the same set of motion sensors $\mathcal{V}$ with an identical adjacency matrix $A$. We want to generalize this problem to tasks with heterogeneous sensors, meaning that the underlying graph structure and the set of vertices vary across tasks ($\mathcal{G}_m\neq \mathcal{G}_{m'} \ \forall m,m'\in M\ (m\neq m')$), while it is shared between support and query data of the same task. Thus, the number of motion sensors $C$ is not fixed and depends on the task at hand.
  
\subsection{GraphHetNet}

\begin{figure}[t] \label{fig:network} \centering
    \includegraphics[width=0.96\textwidth]{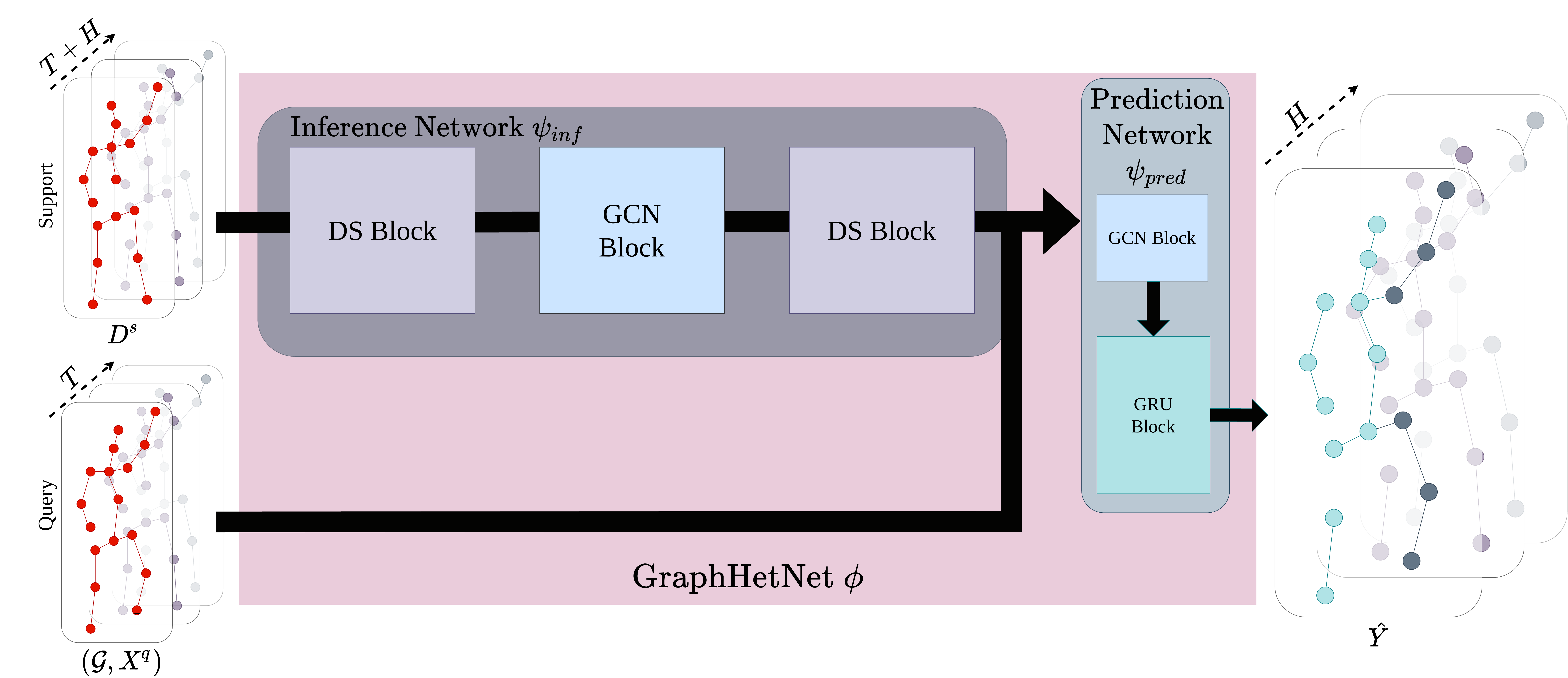}
    \caption{The pipeline for our proposed approach \textsc{GraphHetNet}. \textit{DS Block} stands for Deep Set Block and \textit{GCN Block} for Graph Convolution Network Block. The network takes the full support data $D^s=(\mathcal{G},X^s,Y^s)$ and the predictors of the query data $(\mathcal{G},X^q)$ and outputs a set of outputs $\hat{y}$ represent the next $H$ frames after the $T$ frames of the instances in $D^q$. Batch dimensions are omitted for simplicity.}
\end{figure}
Our model \graphhetnet{} denoted by $\phi$ is based on \netname{} \cite{brinkmeyer2022few}, which uses a set approach for few-shot time-series forecasting with heterogeneous attributes similar to the approach of Iwata et al. \cite{iwata2020meta}. The overall architecture consists of two main components: First, the inference network, which processes the predictor and target data of the support set $D^s$ of a task to generate a latent task representation which should contain useful information to forecast the query instances. Second, the prediction network computes the actual motion forecast for the query set $D^q$ of the task at hand based on its predictors and the task embedding of the support network. In prior approaches \cite{iwata2020meta,brinkmeyer2022few}, both components are composed of multiple stacked deep set blocks (\textit{DS Block}), which process the input data as a set of attributes. To compute the embeddings for every single vertex $c\in C$ over the instances $I$ of the support data $D^s$, a single layer in such a block is then a deep set layer \cite{zaheer2017deep}:
\begin{equation}
    w_c= g_{\text{\textsc{ds}}}\left(\frac{1}{I} \sum_{i=1}^{I} f_{\text{\textsc{ds}}}(x_{ic})\right) \ \forall c\in C
    \label{eq:dsb}
\end{equation}
Here, $w_c\in \mathbb{R}^{T\times K}$ with K being the latent output dimension of $g_{\text{DS}}$. By employing an inner function $f_{\text{\textsc{ds}}}:\mathbb{R}^{T\times 1}\rightarrow \mathbb{R}^{T\times K}$ on each element of the set of instances $X$, and an outer function $g_{\text{\textsc{ds}}}:\mathbb{R}^{T\times K}\rightarrow \mathbb{R}^{T\times K}$ on the aggregation of this set, we can model a permutation-invariant layer that operates on the set of instances. The theoretical foundation of this layer lies in the Kolmogorov-Arnold representation theorem, which states that any multivariate continuous function can be written as a finite composition of continuous functions of a single variable and the binary operation of addition~\cite{schmidt2021kolmogorov}.

In contrast to previous approaches that operate on heterogeneous attributes, we do not utilize \textit{DS} blocks to aggregate the information across attributes, but only across instances, as our problem's attributes are motion sensors structured in a graph and not in a set. Instead, we include blocks of graph convolutional layers (\textit{GCN Block}) \cite{kipf2016semi} in both the inference and the prediction network. We can then aggregate information across sensors by stacking graph convolutional layers. A single layer in the block is then defined as:
\begin{equation}
    u_{ic}= g_{\text{\textsc{gcn}}}\left(\left[x_{ic},\sum_{j\in N(c)} f_{\text{\textsc{gcn}}}(x_{ij})\right]\right) \ \forall c\in C \ \forall i\in I
\end{equation}
where $u_{ic}\in\mathbb{R}^{T\times K}$, $N(c)$ is the set of all vertices that are in the neighborhood of $c$ meaning $A_{cj}=1$ for every $j\in N(c)$ and $[.]$ is the concatenation along the latent feature axis. The inner function $f_{\text{\textsc{gcn}}}:\mathbb{R}^{T\times 1}\rightarrow \mathbb{R}^{T\times K}$ prepares the neighbor embeddings, while the outer function $g_{\text{\textsc{gcn}}}:\mathbb{R}^{T\times 2K}\rightarrow \mathbb{R}^{T\times K}$ updates the vertex features of the respective sensor with its aggregated neighbor messages. Note that this layer only captures the information across motion sensors, not instances. The models $g_{\text{\textsc{gcn}}},f_{\text{\textsc{gcn}}},g_{\text{\textsc{ds}}},f_{\text{\textsc{ds}}}$ are Gated recurrent units (GRU) to deal with the temporal information.
As shown in Figure \ref{fig:network}, our full model \graphhetnet{} $\phi$ consists of the two model components inference and prediction network. The inference network $\psi_{\text{inf}}$ processes the full support data $D^s$ to compute the task embeddings across instances and motion sensors. The prediction network $\psi_{\text{pred}}$ processes the query data to output the final forecast. Thus, the prediction $\hat{Y}$ of our model for a task $m$ is given by:
\begin{equation}
    \hat{Y}_m=\phi(X^q_m,G^q_m,D^s) = \psi_{\text{pred}}(X^q_m,G^q_m,\psi_{\text{inf}}(G^s_m,X^s_m,Y^s_m))
\end{equation}
The inference model $\psi_{\text{inf}}$ is composed of a \textit{GCN} block in between two \textit{DS} blocks to capture both information across instances and motion sensors. The prediction network $\psi_{\text{pred}}$ consists of a \textit{GCN} block, followed by a block of stacked GRU layers (\textit{GRU Block}) which compute the target motion forecast.

     \section{Results} \label{chap:experiments}

We conducted multiple experiments on the Human3.6M dataset \cite{h36m_pami,IonescuSminchisescu11}, consisting of 17 motion categories recorded for 11 subjects, resulting in 3.6 million frames in total. We want to evaluate our approach for few-shot motion tasks with heterogeneous sensors such that each task contains a subset of the vertices of the full motion graph, with the graph of each task being an induced subgraph of the original one. We also conduct an ablation on the standard few-shot motion prediction setting proposed in prior approaches \cite{gui2018few,zang2021few,zang2022few} that considers homogeneous tasks only, meaning each task contains all sensors
in identical order.

\subsection{Experimental setup}
In both cases, we have 11 actions in meta-training (directions, greeting, phoning, posing, purchases, sitting, sitting down, taking a photo, waiting, walking a dog, and walking together) and 4 actions in meta-testing (walking, eating, smoking, and discussion). Furthermore, we also utilize the same split across subjects for meta-test and meta-training as proposed by Gui et al. \cite{gui2018few}. The task is to forecast the next 10 frames (400ms) given the previous 50 frames (2000ms) across the given set of sensors. A single task consists of five support instances and two query instances which means that the model needs to adapt to a previously unseen action based on five labeled instances only. During meta-training, each meta-batch consists of one task per action totaling 11 tasks. The tasks in the classical setting contain all nonzero angles for each of the 32 joints totaling 54 angles as motion sensors. In our main experiment on heterogeneous sensors, each task has only a subset of the set of all sensors. In particular, we sample an induced subgraph of the original human skeleton graph by selecting a random sensor as the initial root node and then recursively adding a subset of neighboring vertices to the graph, including all edges whose endpoints are both in the current subset. The statistics of the original motion graph of Human3.6M and our sampled induced subgraphs are given in Table~\ref{tab:stats}. The number of unique tasks we sample during our experiments is enormous, as is the number of possible induced subgraphs from a given source graph. We evaluated this empirically by sampling one million tasks from the full graph and found around 842,872 unique tasks, meaning only around $16\%$ of the subgraphs were sampled more than once. This guarantees that many tasks our model encounters during meta-testing are previously unseen. More details on the task sampling procedure are stated in the appendix.
\begin{table}[t]
\begin{tabular}{p{2cm}p{0.8cm}p{0.8cm}p{0.8cm}p{0.8cm}p{0.8cm}p{0.4cm}p{0.8cm}p{0.8cm}p{0.8cm}p{0.8cm}p{0.8cm}}
\multicolumn{1}{l|}{}                    &               & walking       &               & \multicolumn{1}{l|}{}              &               & \multicolumn{1}{l|}{} &               & smoking       &               & \multicolumn{1}{l|}{}              &               \\
\multicolumn{1}{l|}{}                    & 80            & 160           & 320           & \multicolumn{1}{l|}{400}           & Avg           & \multicolumn{1}{l|}{} & 80            & 160           & 320           & \multicolumn{1}{l|}{400}           & Avg           \\ \cline{1-6} \cline{8-12} 
\multicolumn{1}{l|}{$\text{\textsc{res-sup}}_{\text{single}}$ \cite{martinez2017human}} & 0.65          & 1.15          & 2.06          & \multicolumn{1}{l|}{2.40}          & 1.57          & \multicolumn{1}{l|}{} & 0.76          & 1.17          & 1.99          & \multicolumn{1}{l|}{2.02}          & 1.48          \\
\multicolumn{1}{l|}{$\text{\textsc{res-sup}}_{\text{all}}$ \cite{martinez2017human}}    & 0.88          & 1.11          & 1.17          & \multicolumn{1}{l|}{1.20}          & 1.09          & \multicolumn{1}{l|}{} & 1.47          & 1.69          & 1.14          & \multicolumn{1}{l|}{1.4}           & 1.43          \\
\multicolumn{1}{l|}{$\text{\textsc{res-sup}}_{\text{trans}}$ \cite{martinez2017human}}            & 0.85          & 1.18          & 1.19          & \multicolumn{1}{l|}{1.17}          & 1.09          & \multicolumn{1}{l|}{} & 1.10          & 1.47          & 1.73          & \multicolumn{1}{l|}{1.94}          & 1.56          \\
\multicolumn{1}{l|}{\textsc{paml} \cite{gui2018few}}       & 0.26          & 0.39          & 0.56          & \multicolumn{1}{l|}{0.64}          & 0.46          & \multicolumn{1}{l|}{} & 0.58          & 0.64          & 0.69          & \multicolumn{1}{l|}{0.83}          & 0.69          \\
\multicolumn{1}{l|}{\textsc{TimeHet} \cite{brinkmeyer2022few}} & {\ul 0.23}          & {\ul 0.30}          & 0.44          & \multicolumn{1}{l|}{0.53}          & {\ul 0.37}          & \multicolumn{1}{l|}{} & {\ul 0.49}          & {\ul 0.52}          & 0.58          & \multicolumn{1}{l|}{0.62}          & 0.55          \\
\multicolumn{1}{l|}{\textsc{MoPred} \cite{zang2021few,zang2022few}} & 0.26        & 0.33    & {\ul 0.43}     &  \multicolumn{1}{l|}{{\ul 0.52}}     & 0.39     & \multicolumn{1}{l|}{} & 0.51        & {\ul 0.52}       & {\ul 0.54}      & \multicolumn{1}{l|}{{\ul 0.61} }        & {\ul 0.54 }      \\
\multicolumn{1}{l|}{\textsc{GHN} (ours)}                & \textbf{0.17} & \textbf{0.22} & \textbf{0.30} & \multicolumn{1}{l|}{\textbf{0.37}} & \textbf{0.27} & \multicolumn{1}{l|}{} & \textbf{0.41} & \textbf{0.42} & \textbf{0.43} & \multicolumn{1}{l|}{\textbf{0.48}} & \textbf{0.44}\\
\cline{1-6} \cline{8-12} 

\multicolumn{1}{l|}{Lift in \%}                & 26.1 & 26.7 & 30.2 & \multicolumn{1}{l|}{28.8} & 27.0 & \multicolumn{1}{l|}{} & 16.3 & 19.2 & 20.4 & \multicolumn{1}{l|}{21.3} & 18.5 \\ 
                                         &               &               &               &                                    &               &                       &               &               &               &                                    &               \\
\multicolumn{1}{l|}{}                    &               & discussion    &               & \multicolumn{1}{l|}{}              &               & \multicolumn{1}{l|}{} &               & eating        &               & \multicolumn{1}{l|}{}              &               \\
\multicolumn{1}{l|}{}                    & 80            & 160           & 320           & \multicolumn{1}{l|}{400}           &               & \multicolumn{1}{l|}{} & 80            & 160           & 320           & \multicolumn{1}{l|}{400}           &               \\ \cline{1-6} \cline{8-12} 
\multicolumn{1}{l|}{$\text{\textsc{res-sup}}_{\text{single}}$ \cite{martinez2017human}} & 0.97          & 1.56          & 1.86          & \multicolumn{1}{l|}{2.67}          & 1.77          & \multicolumn{1}{l|}{} & 0.55          & 0.93          & 1.54          & \multicolumn{1}{l|}{1.74}          & 1.19          \\
\multicolumn{1}{l|}{$\text{\textsc{res-sup}}_{\text{all}}$ \cite{martinez2017human}}    & 0.96          & 1.11          & 1.30          & \multicolumn{1}{l|}{1.44}          & 1.2           & \multicolumn{1}{l|}{} & 0.85          & 1.03          & 0.92          & \multicolumn{1}{l|}{1.05}          & 0.96          \\
\multicolumn{1}{l|}{$\text{\textsc{res-sup}}_{\text{trans}}$ \cite{martinez2017human}}            & 1.30          & 1.42          & 1.68          & \multicolumn{1}{l|}{1.75}          & 1.53          & \multicolumn{1}{l|}{} & 0.68          & 0.78          & 0.94          & \multicolumn{1}{l|}{1.03}          & 0.86          \\
\multicolumn{1}{l|}{PAML \cite{gui2018few}}       & 0.35          & 0.52          & 0.78          & \multicolumn{1}{l|}{0.91}          & 0.64          & \multicolumn{1}{l|}{} & 0.23          & 0.28          & 0.42          & \multicolumn{1}{l|}{0.56}          & 0.37          \\
\multicolumn{1}{l|}{\textsc{TimeHet} \cite{brinkmeyer2022few}} & {\ul 0.29}          & {\ul 0.42}          & 0.69          & \multicolumn{1}{l|}{0.85}          & 0.59          & \multicolumn{1}{l|}{} & {\ul 0.20}           & {\ul 0.28}          & 0.41          & \multicolumn{1}{l|}{{\ul 0.52}}          & {\ul 0.35 }         \\
\multicolumn{1}{l|}{\textsc{MoPred} \cite{zang2021few,zang2022few}} & 0.34        & {\ul 0.42}       & {\ul 0.62}    &  \multicolumn{1}{l|}{{\ul 0.77}}     & {\ul 0.54}     & \multicolumn{1}{l|}{} & 0.25        & 0.29        &    {\ul 0.39}           &  \multicolumn{1}{l|}{0.59}     &     0.38    \\
\multicolumn{1}{l|}{\textsc{GHN} (ours)}                & \textbf{0.22} & \textbf{0.30} & \textbf{0.55} & \multicolumn{1}{l|}{\textbf{0.69}} & \textbf{0.44} & \multicolumn{1}{l|}{} & \textbf{0.14} & \textbf{0.17} & \textbf{0.25} & \multicolumn{1}{l|}{\textbf{0.34}} & \textbf{0.22}\\
\cline{1-6} \cline{8-12} 

\multicolumn{1}{l|}{Lift in \%}                & 24.1 & 28.6 & 11.3 & \multicolumn{1}{l|}{10.4} & 18.5 & \multicolumn{1}{l|}{} & 30.0 & 39.3 & 35.9 & \multicolumn{1}{l|}{34.6} & 37.1
\end{tabular}
\\

\caption{Results few-shot motion prediction with heterogenous sensors given in Mean Angle Error of different methods on Human3.6M. Best results are in bold, second best are underlined. The percentage improvement is given for our model compared to the respective second-best one.}
\label{table:resultsSub}
\end{table}
 
\begin{figure}[t]  \centering
    \includegraphics[width=1\textwidth]{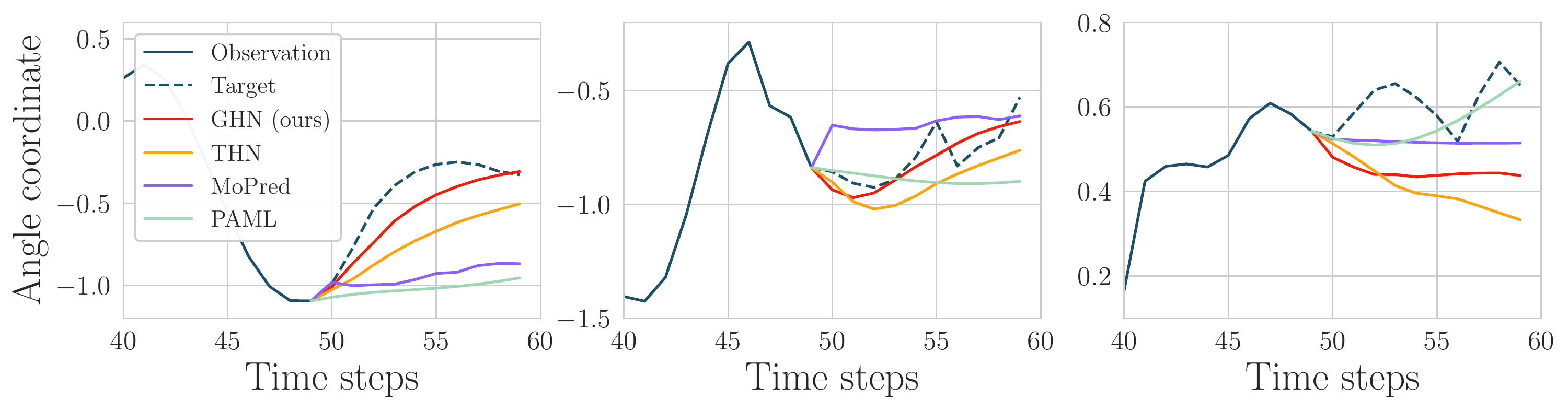}
    \caption{Examples of three motion predictions in exponential map for \ght and baseline approaches. We sampled two examples where our approach (red) has the lowest error and one where a baseline performs best. Full past horizon is shown in appendix.}
    \label{fig:liney}
\end{figure}

We compare against three non-meta-learning baselines, which are variations of the popular detection network \textsc{residual-sup} \cite{martinez2017human}, which consists of stacked GRU's with residual skip connections: $\text{\textsc{res-sup}}_{single}$ trains the model on the support data of the test task at hand only while evaluating the query data. $\text{\textsc{res-sup}}_{all}$ trains the model on the data of all the meta-training actions in standard supervised fashion. In the case of the heterogeneous tasks, the sensor dimension is padded with zeros to 54 since the model is not equipped to deal with heterogeneous sensor sets. The query data of the meta-test tasks is used to evaluate the final performance. $\text{\textsc{res-sup}}_{trans}$ uses $\text{\textsc{res-sup}}_{all}$ as a pretrained model to then fine-tune it to the support data of the test task at hand before evaluating the query data of it.
Furthermore, we compare against the few-shot motion baselines \textsc{paml} \cite{gui2018few}, and \textsc{MoPredNet} \cite{zang2021few,zang2022few}, which both evaluate their approach in the homogeneous setup, as well as \netname{} \cite{brinkmeyer2022few} as it is the first model for time-series forecasting across heterogeneous attributes. Both \textsc{paml} and \textsc{MoPredNet} do not have any publicized code (and we could not reach the authors about it), which is why the results for the standard setting are taken from their respective published results. At the same time, we re-implemented both models to evaluate them on the heterogeneous setup. For \netname{}, we utilize the officially published code. We had to adapt it as the original model is built to forecast a single target variable given a set of covariates that span the future time horizon. In contrast, we want to forecast multiple variables in a set without any given future covariates. The adapted version of \netname{}, as well as the reimplementations of \textsc{paml} and MoPredNet, and the appendix, can also be found in our link: \url{https://github.com/brinkL/graphhetnet}.
We optimized the hyperparameters of all models via grid search. For our approach, the best found configuration includes two graph convolutional layers per \textit{GCN} block, the \textit{DS} blocks contain three stacked GRUs each, and the number of units per GRU is 64. We optimize our model with Adam and a learning rate of 0.0001.

\begin{table}[!t]
    \begin{minipage}{.45\linewidth}
 
      \centering
        \begin{tabular}{c|c|c}
                 & Full      & Sampled     \\ \hline
        vertices         & $54$        & $26.8\pm 12.9$ \\
        edges per vertex & $6.6\pm 3.1$ & $3.9\pm 1.7$  
        \end{tabular}
        \vspace{1em}
        \caption{Statistics of the full Human3.6M graph and for the subgraphs sampled during training on tasks with heterogeneous attributes.}
    \label{tab:stats}
    \end{minipage}\hspace{.09\linewidth}
\begin{minipage}{.45\linewidth}
      \centering
        \begin{tabular}{c|c|c|c|c}
                & PAML & MoPred  & \textsc{THN} & \textsc{GHN} \\ \hline
        Param.           & 3,373K   & 40,945K    & 661K & 265K    \\ 
\end{tabular}
        \\ ~ 

        \caption{Number of parameters in the models PAML, MoPred, \netname{} (as \textsc{THN}), and \graphetnet{} (as \textsc{GHN}, ours) in multiples of 1000.
        }
    \label{tab:sizes}
    \end{minipage} 
\end{table} 
\subsection{Results}
\begin{figure}[t!] \label{network} \centering
    \includegraphics[width=.68\textwidth]{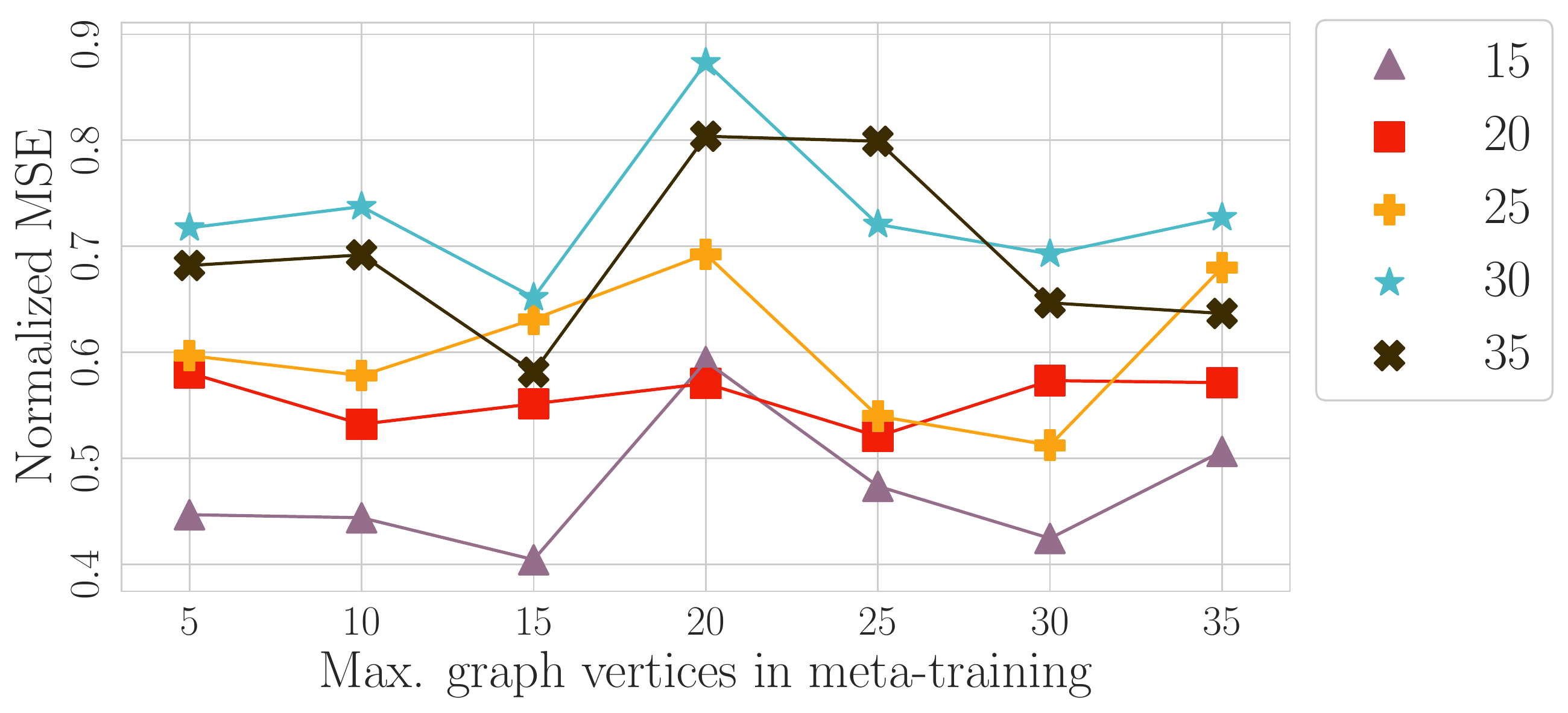}
    \caption{Each line represents a model evaluated for tasks up to a certain number of sensors in test, while the x-axis shows the maximum number of sensors in meta-training. Results are given in MSE averaged across the normalized results for each action.}
    \label{fig:heat}
\end{figure}
The results for our experiment on few-shot tasks with heterogeneous sensors are shown in Table~\ref{table:resultsSub}. Our approach outperforms all baselines with significant margins over all actions and time horizons. The performance improvements compared to the respective second best approach range from 10.4 percent for 400ms on the action "\textit{discussion}" to 39.3 percent for the motion prediction at 160ms for the action "\textit{eating}." The second-best results are shared between \netname{} and \textsc{MoPredNet} (abbreviated \textsc{MoPred} in the table). Three examples for motion forecasts of this experiment are given in the Figure~\ref{fig:liney} for two tasks where our approach has the highest performance and one task where a baseline approach performs better. As expected, the motion prediction of \graphhetnet{} is most similar to \netname{} with our method being more accurate. When comparing the model capacity of our approach and the analyzed baselines based on the model parameters illustrated in Table~\ref{tab:sizes}, one can see that our model contains significantly fewer parameters, with two magnitudes difference to \textsc{MoPredNet} \cite{zang2022few}. \netname{} is the closest with double the number of parameters. Further experimental results for all actions of the Human3.6M can be found in our appendix.

\begin{table}[t]
\begin{tabular}{p{2cm}p{0.8cm}p{0.8cm}p{0.8cm}p{0.8cm}p{0.8cm}p{0.4cm}p{0.8cm}p{0.8cm}p{0.8cm}p{0.8cm}p{0.8cm}}
\multicolumn{1}{l|}{}                    &               & walking       &               & \multicolumn{1}{l|}{}              &               & \multicolumn{1}{l|}{} &               & smoking       &               & \multicolumn{1}{l|}{}              &               \\
\multicolumn{1}{l|}{}                    & 80            & 160           & 320           & \multicolumn{1}{l|}{400}           & Avg           & \multicolumn{1}{l|}{} & 80            & 160           & 320           & \multicolumn{1}{l|}{400}           & Avg           \\ \cline{1-6} \cline{8-12} 
\multicolumn{1}{l|}{$\text{\textsc{res-sup}}_{\text{single}}$ \cite{martinez2017human}$*$} & 0.39          & 0.69          & 0.97          & \multicolumn{1}{l|}{1.08}          & 0.78          & \multicolumn{1}{l|}{} & 0.27          & 0.50          & 0.98          & \multicolumn{1}{l|}{1.00}          & 0.69          \\
\multicolumn{1}{l|}{$\text{\textsc{res-sup}}_{\text{all}}$ \cite{martinez2017human}*}    & 0.36          & 0.61          & 0.84          & \multicolumn{1}{l|}{0.95}          & 0.69          & \multicolumn{1}{l|}{} & {\ul 0.26}    & 0.49          & 0.98          & \multicolumn{1}{l|}{0.97}          & 0.68          \\
\multicolumn{1}{l|}{$\text{\textsc{res-sup}}_{\text{trans}}$ \cite{martinez2017human}*}            & 0.34          & 0.57          & 0.78          & \multicolumn{1}{l|}{{\ul 0.89}}    & 0.65          & \multicolumn{1}{l|}{} & {\ul 0.26}    & 0.48          & 0.93          & \multicolumn{1}{l|}{0.91}          & 0.65          \\
\multicolumn{1}{l|}{\textsc{paml} \cite{gui2018few}*}       & 0.4           & 0.69          & 0.97          & \multicolumn{1}{l|}{1.08}          & 0.79          & \multicolumn{1}{l|}{} & 0.34          & 0.63          & 1.13          & \multicolumn{1}{l|}{1.12}          & 0.80          \\
\multicolumn{1}{l|}{\textsc{TimeHet} \cite{brinkmeyer2022few}} & 0.32          & {\ul 0.37}    & 0.70          & \multicolumn{1}{l|}{0.94}          & 0.58          & \multicolumn{1}{l|}{} & 0.43          & 0.46          & {\ul 0.69}    & \multicolumn{1}{l|}{{\ul 0.68}}    & {\ul 0.57}    \\
\multicolumn{1}{l|}{\textsc{MoPred} (reimp.)}          & 0.42          & 0.52          & 0.77          & \multicolumn{1}{l|}{0.98}          & 0.67          & \multicolumn{1}{l|}{} & 0.48          & 0.54          & 0.71          & \multicolumn{1}{l|}{0.94}          & 0.67          \\
\multicolumn{1}{l|}{\textsc{MoPred} \cite{zang2021few,zang2022few}*} & {\ul 0.21}    & \textbf{0.35} & \textbf{0.55} & \multicolumn{1}{l|}{\textbf{0.69}} & \textbf{0.45} & \multicolumn{1}{l|}{} & {\ul 0.26}    & {\ul 0.47}    & 0.93          & \multicolumn{1}{l|}{0.9}           & 0.64          \\
\multicolumn{1}{l|}{\textsc{ghn} (ours)}                & \textbf{0.17} & \textbf{0.35} & {\ul 0.69}    & \multicolumn{1}{l|}{0.94}          & {\ul 0.54}    & \multicolumn{1}{l|}{} & \textbf{0.12} & \textbf{0.17} & \textbf{0.67} & \multicolumn{1}{l|}{\textbf{0.54}} & \textbf{0.38} \\ 
\cline{1-6} \cline{8-12} 

\multicolumn{1}{l|}{Lift in \%}                & 19.0 & 0.0 & -25.5 & \multicolumn{1}{l|}{-36.2} & -20.0 & \multicolumn{1}{l|}{} & 53.8 & 63.8 & 28.0 & \multicolumn{1}{l|}{40.0} & 40.6 \\ 
                                         &               &               &               &                                    &               &                       &               &               &               &                                    &               \\
\multicolumn{1}{l|}{}                    &               & discussion    &               & \multicolumn{1}{l|}{}              &               & \multicolumn{1}{l|}{} &               & eating        &               & \multicolumn{1}{l|}{}              &               \\
\multicolumn{1}{l|}{Horizon in ms}                    & 80            & 160           & 320           & \multicolumn{1}{l|}{400}           &               & \multicolumn{1}{l|}{} & 80            & 160           & 320           & \multicolumn{1}{l|}{400}           &               \\ \cline{1-6} \cline{8-12} 
\multicolumn{1}{l|}{$\text{\textsc{res-sup}}_{\text{single}}$ \cite{martinez2017human}*} & 0.32          & 0.66          & 0.95          & \multicolumn{1}{l|}{1.09}          & 0.76          & \multicolumn{1}{l|}{} & 0.28          & 0.50          & 0.77          & \multicolumn{1}{l|}{0.91}          & 0.62          \\
\multicolumn{1}{l|}{$\text{\textsc{res-sup}}_{\text{all}}$ \cite{martinez2017human}$*$}    & 0.31          & 0.66          & 0.94          & \multicolumn{1}{l|}{{\ul 1.03}}    & 0.74          & \multicolumn{1}{l|}{} & 0.26          & 0.46          & 0.70          & \multicolumn{1}{l|}{0.82}          & 0.56          \\
\multicolumn{1}{l|}{$\text{\textsc{res-sup}}_{\text{trans}}$ \cite{martinez2017human}$*$}            & 0.30          & 0.65          & {\ul 0.91}    & \multicolumn{1}{l|}{0.99}          & 0.71          & \multicolumn{1}{l|}{} & 0.22          & 0.35          & 0.54          & \multicolumn{1}{l|}{\textbf{0.69}} & 0.45          \\
\multicolumn{1}{l|}{\textsc{paml} \cite{gui2018few}*}       & 0.36          & 0.72          & 1.03          & \multicolumn{1}{l|}{1.15}          & 0.82          & \multicolumn{1}{l|}{} & 0.29          & 0.51          & 0.8           & \multicolumn{1}{l|}{0.95}          & 0.64          \\
\multicolumn{1}{l|}{\textsc{TimeHet} \cite{brinkmeyer2022few}} & 0.33          & {\ul 0.49}    & 1.00          & \multicolumn{1}{l|}{1.31}          & 0.78          & \multicolumn{1}{l|}{} & 0.28          & 0.35          & 0.61          & \multicolumn{1}{l|}{0.91}          & 0.54          \\
\multicolumn{1}{l|}{\textsc{MoPred} (reimp.)}          & 0.51          & 0.67          & 0.99          & \multicolumn{1}{l|}{1.12}          & 0.82          & \multicolumn{1}{l|}{} & 0.35          & 0.47          & 0.62          & \multicolumn{1}{l|}{0.83}          & 0.56          \\
\multicolumn{1}{l|}{\textsc{MoPred} \cite{zang2021few,zang2022few}*} & {\ul 0.29}    & 0.63          & \textbf{0.89} & \multicolumn{1}{l|}{\textbf{0.98}} & \textbf{0.70} & \multicolumn{1}{l|}{} & {\ul 0.21}    & {\ul 0.34}    & {\ul 0.53}    & \multicolumn{1}{l|}{\textbf{0.69}} & {\ul 0.44}    \\
\multicolumn{1}{l|}{\textsc{ghn} (ours)}                & \textbf{0.19} & \textbf{0.42} & 0.94          & \multicolumn{1}{l|}{1.25}          & \textbf{0.70} & \multicolumn{1}{l|}{} & \textbf{0.17} & \textbf{0.29} & \textbf{0.52} & \multicolumn{1}{l|}{{\ul 0.75}}    & \textbf{0.43}\\
\cline{1-6} \cline{8-12} 

\multicolumn{1}{l|}{Lift in \%}                & 34.5 & 33.3 & -5.6 & \multicolumn{1}{l|}{-27.6} & 0.0 & \multicolumn{1}{l|}{} & 19.0 & 14.7 & 1.9 & \multicolumn{1}{l|}{-8.7} & 2.3
\end{tabular}
\\

\caption{Ablation on homogeneous setting: Mean Angle Error of different methods on Human3.6M dataset for standard few-shot motion prediction task with fixed attribute space. The results with * are taken from the published results of Zang et al. \cite{zang2022few}.}
\label{table:resultsFull}
\end{table} 
\subsection{Ablations}
We also evaluated our model in the standard homogeneous setting where all tasks share a fixed motion graph. This serves the purpose of evaluating whether our model, which is designed for heterogeneous tasks, shows any performance degradation or can perform on par with state-of-the-art approaches in the classical setup.
The results are stated in Table \ref{table:resultsFull}. We implemented our own version of \textsc{paml} and \textsc{MoPredNet}, as there is no public implementation in either approach. We received no further information when contacting the original authors. Both the published results and the results for our implementation of \textsc{MoPredNet} are given in the table, as we were not able to replicate the results reported in the publication. For \textsc{paml}, we only show the reported results as our reimplementation achieves results that match the reported results.
Our approach is shown to be on par with our baselines, \textsc{MoPredNet} while showing slight improvements for short-term frames after 80 and 160 ms. At the same time, the model capacity of our model is two magnitudes lower than of \textsc{MoPredNet} and one lower than \textsc{paml}. Comparing our results to \netname{}, we see that convolutional graph layers give significant performance lifts.  
In a further ablation, we analyzed the influence of the size of sampled subgraphs during meta-training on meta-testing. For this, we repeated our experimental setup but limited the maximum number of nodes in the subgraph from 5 to 35 for meta-training and -testing, respectively. The results in Figure \ref{fig:heat} indicate that our approach is robust to subgraph size in meta-training, with a slight peak when training on tasks up to 20 vertices, demonstrating the model's ability to generalize to larger graphs during testing. This shows how larger tasks correlate to a more difficult motion prediction as the chance to extract useful data from neighbor sensors increases.

     \section{Conclusion} \label{chap:conclusion}

In this work, we proposed a new approach for few-shot human motion prediction, which generalizes over tasks with heterogeneous motion sensors arranged in a graph, outperforming all related baselines which are not equipped for varying sensor graphs. This is the first approach that allows for the prediction of novel human motion tasks independent of their number of sensors. Moreover, using this model, we can rival state-of-the-art approaches for the standard few-shot motion benchmark on tasks with homogeneous sensors while maintaining a significantly smaller model size which can be crucial for applications of human motion detection as these are often found in mobile and handheld devices. By publicizing all our code, including the baselines reimplementation as well as our benchmark pipeline, we hope to motivate future research in this area. 
    \bibliographystyle{splncs04}
    \bibliography{bibt}
    
\newpage
\section*{Appendix}

\subsection*{A \hspace{0.5em} More experimental details}

To sample a heterogeneous few-shot task for a specific action, we first select an induced subgraph of the original human skeleton, as described in Section 4.1. Each task consists of five labeled support instances for training and two query instances for testing.
During meta-testing, all instances are randomly sampled from actor 5. During meta-training, the instances are sampled randomly from the remaining actors, as in previous works \cite{gui2018few,zang2021few,zang2022few}. We train the models for 500 epochs, with 50 meta-batches sampled per epoch. Each meta-batch consists of one task per meta-training action, resulting in a total of 275,000 tasks seen during meta-training. For evaluation, we randomly sample 500 tasks per meta-test action from actor 5.

\subsection*{B \hspace{0.5em} Cross-validation}
We conducted 5-fold cross-validation across actions to assess our model's performance for all actions shown in Table 5. The splits were designed to be intuitively more difficult by grouping similar actions in the same fold, such as walking, walking dog, and walking together. As expected, the results for walking degraded compared to the previous experimental setup proposed by \cite{martinez2017human}.
\begin{table}[]
\renewcommand{\arraystretch}{1.0}
\centering
\smallskip
\begin{tabular}{llllll}
\multicolumn{1}{l|}{}                  & \multicolumn{4}{c|}{GrapHetNet}                   &      \\
\multicolumn{1}{l|}{Actions (by fold)} & 80    & 160   & 320   & \multicolumn{1}{l|}{400}  & Avg  \\ \hline
\multicolumn{1}{l|}{walkingdog}        & 0.409 \hspace{1em} & 0.747 \hspace{1em} & 1.186 \hspace{1em} & \multicolumn{1}{l|}{1.38 \hspace{1em}} & 0.93 \\
\multicolumn{1}{l|}{walkingtogether}   & 0.251 & 0.50  & 0.74  & \multicolumn{1}{l|}{0.82} & 0.58 \\
\multicolumn{1}{l|}{walking}           & 0.17  & 0.32  & 0.50  & \multicolumn{1}{l|}{0.69} & 0.42 \\ \hline
\multicolumn{1}{l|}{smoking}           & 0.233 & 0.23  & 0.51  & \multicolumn{1}{l|}{0.64} & 0.40 \\
\multicolumn{1}{l|}{eating}            & 0.209 & 0.28  & 0.60  & \multicolumn{1}{l|}{0.71} & 0.45 \\
\multicolumn{1}{l|}{posing}            & 0.212 & 0.28  & 0.49  & \multicolumn{1}{l|}{0.67} & 0.41 \\ \hline
\multicolumn{1}{l|}{phoning}           & 1.26  & 2.53  & 2.83  & \multicolumn{1}{l|}{2.98} & 2.40 \\
\multicolumn{1}{l|}{directions}        & 0.61  & 1.12  & 1.43  & \multicolumn{1}{l|}{1.5}  & 1.17 \\
\multicolumn{1}{l|}{waiting}           & 0.257 & 0.39  & 0.77  & \multicolumn{1}{l|}{0.90} & 0.58 \\ \hline
\multicolumn{1}{l|}{discussion}        & 0.19  & 0.46  & 0.97  & \multicolumn{1}{l|}{1.25} & 0.72 \\
\multicolumn{1}{l|}{takingphoto}       & 0.21  & 0.57  & 0.98  & \multicolumn{1}{l|}{1.20} & 0.74 \\
\multicolumn{1}{l|}{sittingdown}       & 0.24  & 0.66  & 1.02  & \multicolumn{1}{l|}{1.19} & 0.78 \\ \hline
\multicolumn{1}{l|}{greeting}          & 0.343 & 0.71  & 1.33  & \multicolumn{1}{l|}{1.71} & 1.02 \\
\multicolumn{1}{l|}{purchases}         & 0.35  & 0.49  & 0.85  & \multicolumn{1}{l|}{0.93} & 0.65 \\
\multicolumn{1}{l|}{sitting}           & 0.32  & 0.40  & 0.67  & \multicolumn{1}{l|}{0.67} & 0.52 \\
\end{tabular}
\caption{Results for 5-fold cross-validation for \graphhetnet{}. Folds are separated by horizontal line.}
\label{tab:cross}
\end{table}
\newpage
\subsection*{C \hspace{0.5em} Forecast with full past horizon} 
    \begin{figure}[h!] \label{fig:fullpred} \centering
        \includegraphics[width=0.81\textwidth]{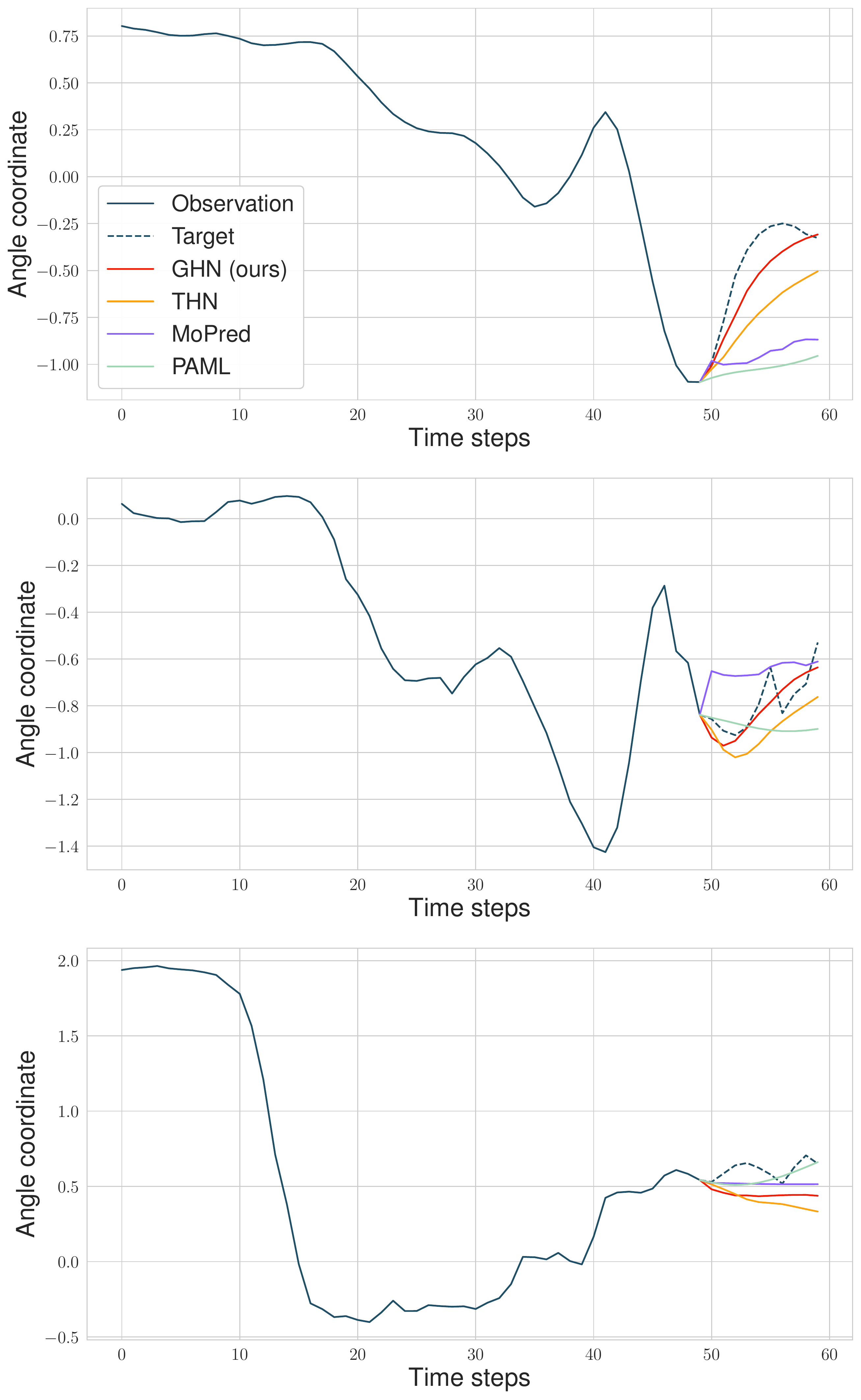}
        \caption{A full zoomed out version of Figure \ref{fig:pred}: Examples of three motion predictions in the exponential map for \ght and baseline approaches. We sampled two examples where our approach (red) has the lowest error and one where a baseline performs best.}
    \end{figure}
\end{document}